\documentclass[10pt,twocolumn,letterpaper]{article}

\pdfoutput=1

\usepackage{iccv}
\usepackage{times}
\usepackage{epsfig}
\usepackage{graphicx}
\usepackage{amsmath}
\usepackage{amssymb}

\usepackage{textcomp}
\usepackage{multirow}
\usepackage{tabularx}
\usepackage{amstext}
\usepackage{threeparttable}

\usepackage[marginal]{footmisc}

\usepackage[breaklinks=true,bookmarks=false]{hyperref}

\iccvfinalcopy 

\def\iccvPaperID{8275} 

\ificcvfinal\fi

\begin{document}

\title{Temporal-wise Attention Spiking Neural Networks for Event Streams Classification}

\author{Man Yao$^{*1}$, Huanhuan Gao$^{*1}$, Guangshe Zhao$^{\dag1}$, 
Dingheng Wang$^{1}$, Yihan Lin$^{2}$, Zhaoxu Yang$^{3}$, Guoqi Li$^{\dag2}$\\
$^1$School of Automation Science and Engineering, Xi’an Jiaotong University, Xi’an 710000, China \\
$^2$Department of Precision Instrument, Center for Brain Inspired Computing Research,\\ Tsinghua University, Beijing 100084, China\\
$^3$School of Aerospace Engineering, Xi’an Jiaotong University, Xi’an 710000, China\\
}




\maketitle
\ificcvfinal\thispagestyle{empty}\fi

\begin{abstract}
How to effectively and efficiently deal with spatio-temporal event streams, where the events are generally sparse and non-uniform and have the µs temporal resolution, is of great value and has various real-life applications. Spiking neural network (SNN), as one of the brain-inspired event-triggered computing models, has the potential to extract effective spatio-temporal features from the event streams. However, when aggregating individual events into frames with a new higher temporal resolution, existing SNN models do not attach importance to that the serial frames have different signal-to-noise ratios since event streams are sparse and non-uniform. This situation interferes with the performance of existing SNNs. In this work, we propose a temporal-wise attention SNN (TA-SNN) model to learn frame-based representation for processing event streams. Concretely, we extend the attention concept to temporal-wise input to judge the significance of frames for the final decision at the training stage, and discard the irrelevant frames at the inference stage. We demonstrate that TA-SNN models improve the accuracy of event streams classification tasks. We also study the impact of multiple-scale temporal resolutions for frame-based representation. Our approach is tested on three different classification tasks: gesture recognition, image classification, and spoken digit recognition. We report the state-of-the-art results on these tasks, and get the essential improvement of accuracy (almost 19\%) for gesture recognition with only 60 ms.  
\end{abstract}
\footnote{$^*$ These authors contribute equally to this work.}
\footnote{$^\dag$ Corresponding authors, liguoqi@tsinghua.edu.cn,\\ zhaogs@mail.xjtu.edu.cn.}

\begin{figure}[!t]
\centering
\includegraphics[scale=0.22]{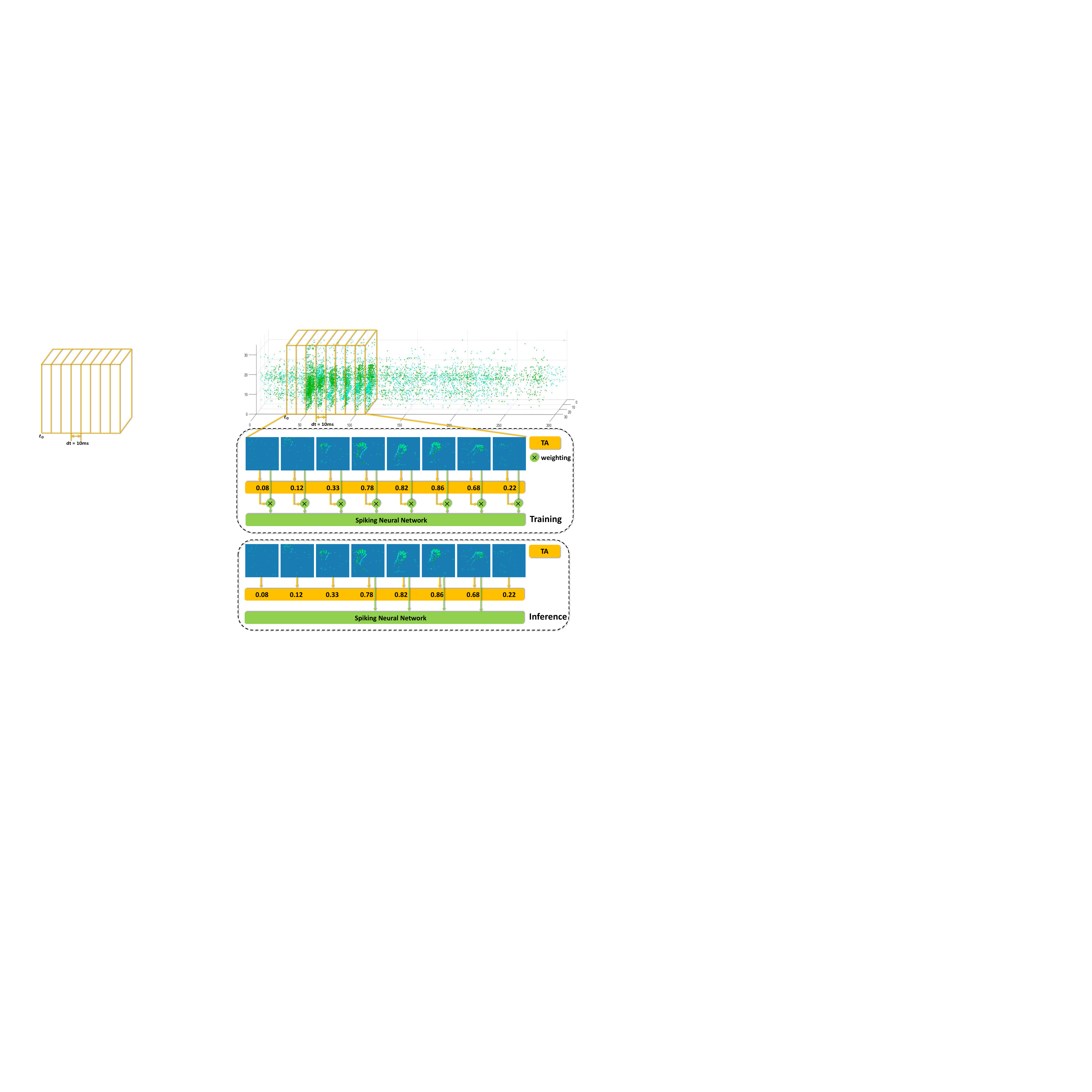}
\caption{Our proposed model use the TA to judge the significance of frames at the training stage and discard the irrelevant frames at the inference stage. The sample is from the DVS128 Gesture dataset. The green and cyan colors denote the On and Off channels which correspond to brightness increase and decrease, respectively (more details of event streams in section \ref{event_data}).}
\label{fig:Event_data_fig}
\end{figure}

\section{Introduction}



Dynamic vision sensors (DVS)\cite{DVS_sensor_1,DVS_sensor_2} pose a new paradigm shift by using sparse and asynchronous events to represent visual information. Unlike the conventional cameras, which produce fixed low-rate synchronized frames (typically less than 60 frames per second), DVS cameras encode the time, location, and polarity of the brightness changes for each pixel at an extremely high event rate (1M to 1G events per second ), and exhibit advantages mainly in three aspects\cite{DVS_Survey,High_Speed_Event_Camera_2}. Firstly, DVS cameras require much less resource, as the events are sparse and only triggered when the intensity changes. Secondly, the \textmu s temporal resolution (TR) of DVS can avoid motion blur by producing high-rate events. Thirdly, DVS cameras have a high dynamic range (140dB vs. 60dB of conventional cameras), which makes them able to acquire information from challenging illumination conditions. These characteristics bring superiorities over conventional cameras when orienting to visual tasks which need low latency, low power consumption, and stability for variant illumination, which have been used in high-speed object tracking\cite{High_Speed_Event_Camera_2}, autonomous driving\cite{Auto_driving_1}, SLAM\cite{Robotic_1}, low-latency interaction\cite{Gesture_Recognition_System}, etc. 

However, we have observed that the event streams recorded by DVS cameras are usually redundant in the temporal dimension, which is caused by high TR and irregular dynamic scene changes. This characteristic makes event streams almost impossible to process directly through deep neural networks (DNNs), which are based on dense computation. Compromising on this, additional data preprocessing\cite{GCN_for_NV_1,hats,DVS_Survey} is required and inevitably dilutes the advantages of low-latency and power-saving of events. Inspired by the working pattern of the mammalian visual cortex\cite{DVS_Survey}, spiking neural networks (SNNs) have a unique event-triggered computation characteristic that can respond to the events in a nearly latency-free and power-saving way\cite{Nature_2,Nature_1,DVS_Survey}, and it is naturally fit for processing event. However, due to the lack of training technology, the performance of deep SNNs has become the biggest obstacle to their application. During experiments, we find that there is still a lot of optimizing room for SNNs to process the event data more efficiently and effectively. That's why we introduce the attention mechanism into SNNs. 

In this work, we propose the temporal-wise attention SNNs (TA-SNNs) by extending the attention concept to temporal-wise input to automatically filter out the irrelevant frames for the final decision. For TA-SNNs, how to implement the attention mechanism while retaining the event-triggered characteristic is the primary consideration. Classic attention methods, such as self-attention\cite{Attention_is_All_You_Need}, are hard to use because the change of network connection destroys the event-triggered characteristic in SNNs. Inspired by squeeze-and-excitation (SE) block\cite{SE_PAMI}, we design the TA module to obtain the statistical features of events at different times, generate the attention scores and then weigh the events by the scores. At the same time, we propose a data augmentation method called \emph{random consecutive slice (RCS)} to utilize the event data. In order to keep the event-encoded data characteristics, we then use binary attention scores at the inference stage with a threshold in the TA module, which is termed as \emph{input attention pruning (IAP)} and obtains an unchanged or even higher accuracy with RCS. Without losing generality, we test our approach with two kinds of SNN models, i.e., leaky integrate-and-fire (LIF) and leaky integrate-and-analog-fire (LIAF), on three kinds of tasks: gesture recognition, image classification, and spoken digit recognition. We report the state-of-the-art results on these tasks in long-term TRs, and get the essential improvement of accuracy (almost 19\%) for gesture recognition with low-latency and power-saving property.

We summarize our contributions as follows:
\begin{itemize}
\item [1)] 
We propose the TA-SNNs for event streams that can undertake the end-to-end training and inference tasks with low latency, low power consumption, and high performance. To the best of our knowledge, this is the first work to introduce temporal-wise attention into SNNs. 
\item [2)] 
We propose the IAP method for SNNs and get similar or even better performance compared with those using full inputs (see Fig.\ref{fig:figure4}). The IAP brings a crucial power-saving significance for SNNs and other event-based algorithms.
\item [3)]
Inspired by the data augmentation method in video recognition\cite{LTC} and overlap method for event stream process\cite{Gesture_Loihi}, we introduce the RCS method to make full use of the sampled data.

\end{itemize}

\section{Related Works}

\textbf{Event Streams Classification}. To yield sufficient signal-to-noise ratios (SNR) for the task accuracy, processing the events as groups is the most common method\cite{DVS_Survey}. In this paper, we adopt the frame-based representation that aggregates event streams into frames\cite{Event_frame_PAMI}. The frame-based representation is easy to generate and naturally compatible with the traditional computer vision framework, and the SNN algorithms based on frames can be easily mapped to neuromorphic hardware\cite{Event_frame_PAMI}. TR is a crucial parameter for frame-based representation, and generally, the bigger TR is, the higher SNR we could have. Most related works are dedicated to using various techniques to improve the classification performance based on the long-term TR, such as improve training method\cite{Going_Deeper_SNN,Direct_training}, change the connection path of the SNNs\cite{LISNN,Csnn}, and hybrid fusion\cite{LIAF,rethink_ann_snn,efficient_snn}, etc. 

\textbf{Spiking Neural Networks}. Spiking neurons, such as the LIF\cite{STBP_Alogorithm}, use spike stream as the data transmission form and connect each other hierarchically as a network, i.e., SNNs. One common way in these spike-based SNNs is to assume that the neurons which have not received any input spikes will skip computations, i.e., event-triggered characteristic\cite{DVS_Survey}. So spike-based SNNs can extract information from spikes in a power-saving way. The other kind of SNNs, i.e., analog-based SNNs, use dynamic characteristics in spiking neurons but transmit analog values in the network, such as LIAF\cite{LIAF}, RELU SRNN\cite{SHD_dataset}, SpArNet\cite{SpArNet}, etc. Analog value makes the network easy to train, but it loses the attributes of skipping computation in spike-based SNNs. Without loss of generality, we separately adopt LIF and LIAF models as the elements of spike-based and analog-based SNNs to test the attention mechanism.  

\textbf{Attention Models}. The attention mechanism selectively focuses on the most informative components of the input and can be interpreted as the sensitivity of the output to the variant input\cite{Attention_2}. The models using attention have been applied to many tasks, such as sequence learning\cite{TAGM,squence_attention_2}, machine translation\cite{Attention_is_All_You_Need,BERT,Non-local}, action recognition\cite{action_attention,action_attention_2}, etc. Generally, there are two types of works , i.e., temporal-wise attention in RNNs\cite{Attention_2} and spatial-wise attention in SNNs\cite{Attention_for_DVS_data,Attention_temporal_encode_SNN,SNN_attention} may be related to the proposed method in this paper. Our work is different from prior works, and we focus on the statistical characteristics of the frames input at different timesteps based on SNNs.

\begin{figure*}[!t]
\centering
\includegraphics[scale=0.3]{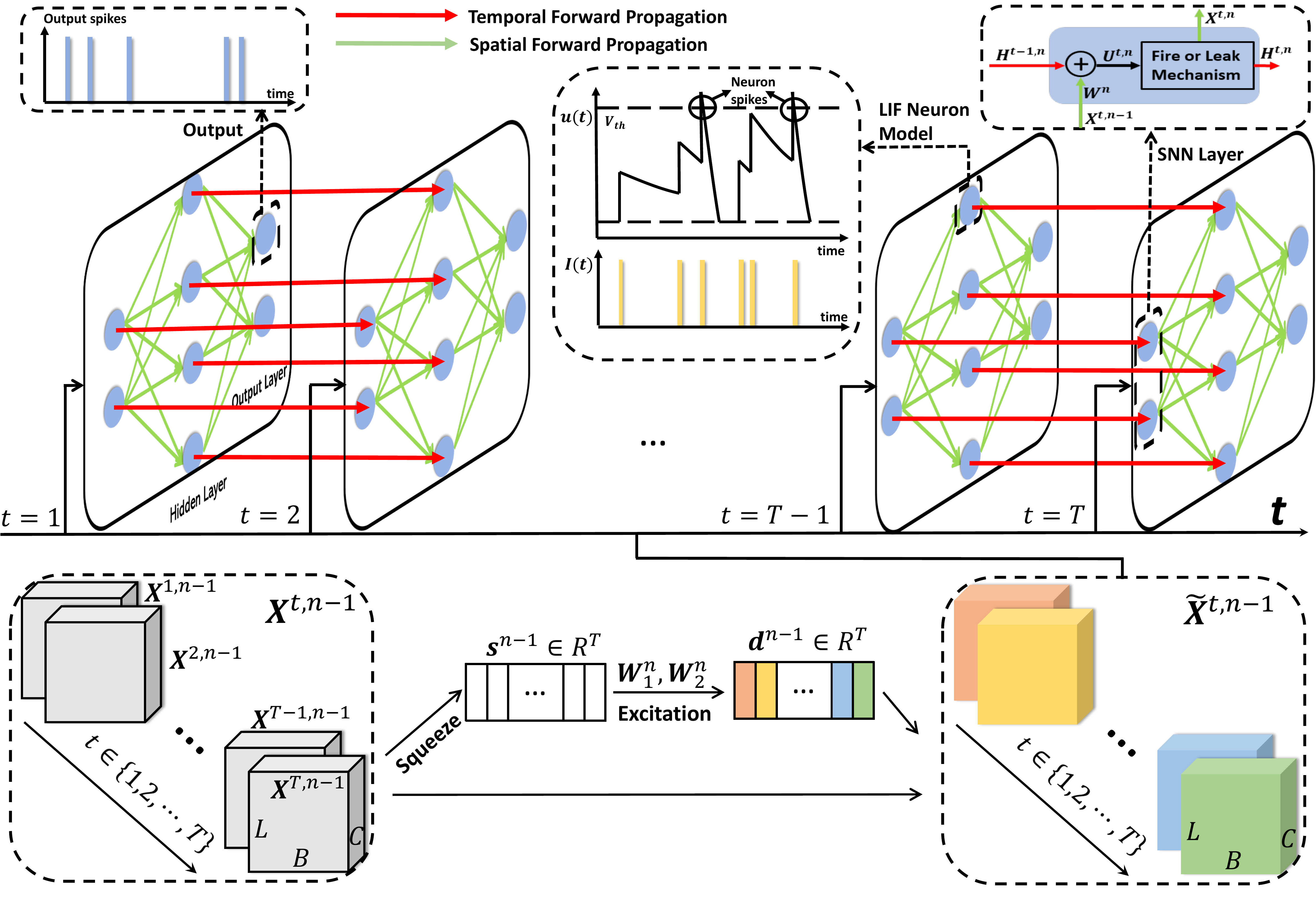}
\caption{Temporal-wise attention spiking neural networks. In score vector $\boldsymbol{d}^{n-1}$, different colors represent different attention scores at different timesteps, multiplying them can produce the new input tensor according to Eq.\ref{Eq:attention_eq3}.}
\label{fig:SNN_Network}
\end{figure*}

\section{Model Description}

\subsection{Frame-based Representation}\label{event_data}
Event steam comprises four dimensions: two spatial coordinates, the timestamp, and the polarity of the event. The polarity indicates an increase (ON) or decrease (OFF) of brightness, where ON/OFF can be represented via +1/-1 values. Assume the TR of event stream is $dt^\prime$ and the spatial resolution is $L\times B$, then the spike pattern tensor $\boldsymbol{X}_{t^\prime}\in\boldsymbol{R}^{L\times B \times 2}$ is equal to events set $E_{t^\prime}=\left\{e_i|e_i=\left[x_i,y_i,t^\prime,p_i\right]\right\}$ at timestamp $t^\prime$. For frame, set a new TR $dt={dt^\prime}\times \beta$, and the consecutive $\beta$ spike patterns can be grouped as a set
\begin{equation}
    E_{t}=\left\{\boldsymbol{X}_{t^\prime}\right\}
    \label{eq:spike_pattern_set}
\end{equation}
where ${t^\prime}\in\left[\beta\times t,\beta\times\left(t+1\right)-1\right]$ and $\beta$ is called resolution factor. Then, the frame of input layer at $t$ time $\boldsymbol{X}^{t, 0}\in\boldsymbol{R}^{L\times B \times 2}$ based on $dt$ can be got by
\begin{equation}
    \boldsymbol{X}^{t, 0} = q(E_{t})
    \label{eq:event_frame_slice}
\end{equation}
where $t\in\left\{1,2,\cdots,T\right\}$ is timestep, and aggregation function $q(\cdot)$ could be selected, as non-polarity aggregation\cite{Gesture_Loihi}, accumulate aggregation\cite{rethink_ann_snn}, AND logic operation aggregation\cite{Compare_SNN_and_RNN}, etc. Here we choose a simple approach, which accumulates event stream with the information of event polarity.

\subsection{Spiking Neural Network Models}
The LIF model is a trade-off between the complex dynamic characteristics of biological neurons and the simpler mathematical form. It is suitable for simulating large-scale SNNs and can be described by a differential function\cite{Nature_2}
\begin{equation}
    \tau\frac{du\left(t\right)}{dt}=-u\left(t\right)+I\left(t\right)  \label{eq:continuous LIF model}
\end{equation}
where $\tau$ is a time constant, and $u\left(t\right)$ and $I\left(t\right)$ are the membrane potential of the postsynaptic neuron and the input collected from presynaptic neurons, respectively (see the relationship of $u\left(t\right)$ and  $I\left(t\right)$ in Fig.\ref{fig:SNN_Network}). For easy inference and training, a simple iterative representation of LIF model\cite{SG_Algorithm} or LIAF model\cite{LIAF} can be described as
\begin{equation}
    \left\{\begin{array}{l}
    \boldsymbol{U}^{t, n}=\boldsymbol{H}^{t-1, n}+g\left(\boldsymbol{W}^{n}, \boldsymbol{X}^{t, n-1}\right) \\
    \boldsymbol{Z}^{t, n}=f\left(\boldsymbol{U}^{t, n}-u_{t h}\right) \\
    \boldsymbol{H}^{t, n}=\left(e^{-\frac{d t}{\tau}} \boldsymbol{U}^{t, n}\right)  \circ\left(\mathbf{1}-\boldsymbol{Z}^{t, n}\right) \\
    \boldsymbol{X}^{t, n}=
    \left\{\begin{array}{l}
    \boldsymbol{Z}^{t, n} \qquad  \qquad  \qquad {\rm for\;LIF}, \\
    ReLU\left(\boldsymbol{U}^{t, n}\right)  \qquad {\rm for\;LIAF},\\
    \end{array}\right. \\
    \end{array}\right. \label{eq:SNN layer}
\end{equation}
where $n$ and $t$ are indices of layer and timestep, $\boldsymbol{W}^{n}$ is the synaptic weight matrix between two adjacent layers, $g(\cdot)$ is a function stands for convolutional operation or fully connected operation, $f(\cdot)$ is a Heaviside step function that satisfies $f\left(x\right)=1$ when $x\geq0$, otherwise $f\left(x\right)=0$, $u_{th}$ is the membrane potential threshold, $e^{-\frac{d t}{\tau}}$ reflects the leakage factor of the membrane potential, $\circ$ is the Hadamard product, $\boldsymbol{X}$ and $\boldsymbol{H}$ are spatial and temporal input, respectively, $\boldsymbol{U}$ is the membrane potential, and $\boldsymbol{Z}$ is the spike tensor. For spatial input tensor $\boldsymbol{X}$, its representation is different for LIF and LIAF which are separately described below.

\textbf{LIF-SNNs}. As shown in Eq.\ref{eq:SNN layer} and Fig.\ref{fig:SNN_Network}, by coupling $\boldsymbol{X}^{t, n-1}$ from the $n-1$ layer and $\boldsymbol{H}^{t-1, n}$ from the $t-1$ timestep, we can get $\boldsymbol{U}^{t, n}$. If $\boldsymbol{U}^{t, n}$ is greater than $u_{th}$, the neuron executes the fire mechanism, which outputs $\boldsymbol{Z}^{t, n}$ as the spatial input of next layer, i.e., $\boldsymbol{X}^{t, n}=\boldsymbol{Z}^{t, n}$, and resets $\boldsymbol{U}^{t, n}$ to $u_{rest}$. Meanwhile, the neuron executes the leak mechanism, and the decayed value of membrane potential $\boldsymbol{H}^{t, n}$ will be used as the temporal input for the next timestep.

\textbf{LIAF-SNNs}. For the LIAF, it keeps the $\boldsymbol{H}^{t-1, n}$ and changes the Heaviside step function to ReLU function for $\boldsymbol{U}^{t, n}$, i.e., $\boldsymbol{X}^{t, n}=ReLU\left(\boldsymbol{U}^{t, n}\right)$, then both spatial and temporal domains are analog values. We use the STBP\cite{STBP_Alogorithm} and the BPTT algorithm\cite{LIAF} to train LIF-SNNs and LIAF-SNNs, respectively.

\subsection{Temporal-wise Attention for SNNs}

The goal of TA module is to estimate the saliency of each frame. This saliency score should not only be based on the input statistical characteristic at the current timestep, but also take into consideration the information from neighboring frames. We apply the squeeze step and excitation step\cite{SE_PAMI} in temporal-wise to implement the above two points. The spatial input tensor of $n$th layer at $t$th timestep is $\boldsymbol{X}^{t,n-1}\in\boldsymbol{R}^{L\times B\times C}$ where $C$ is channel size. 

Squeeze step calculates a statistical vector of event numbers, and the value of statistical vector $\boldsymbol{s}^{n-1}\in\boldsymbol{R}^{T}$ at $t$th timestep is
\begin{equation}
\boldsymbol{s}^{n-1}_{t}=\frac{1}{L \times B \times C} \sum_{k=1}^{C} \sum_{i=1}^{L} \sum_{j=1}^{B} \boldsymbol{X}^{t,n-1}(k, i, j) .
\label{Eq:attention_eq1}
\end{equation}
By executing the excitation step, $\boldsymbol{s}^{n-1}$ is subjected to nonlinear mapping through a two-layer fully connected network to obtain the correlation between different frames, i.e., score vector
\begin{equation}
    \boldsymbol{d}^{n-1}=
    \left\{\begin{array}{l}
    \sigma\left(\boldsymbol{W}^{n}_{2} \delta\left(\boldsymbol{W}^{n}_{1} \boldsymbol{s}^{n-1}\right)\right) \;\; \qquad \qquad \ {\rm training}, \\
    f\left(\sigma\left(\boldsymbol{W}^{n}_{2} \delta\left(\boldsymbol{W}^{n}_{1} \boldsymbol{s}^{n-1}\right)\right)-d_{t h}\right)  \;\; {\rm inference},\\
    \end{array}\right. 
    \label{Eq:traning_and_inference}
\end{equation}
where $\delta$ and $\sigma$ are ReLU and sigmoid activation function, respectively, $\boldsymbol{W}^{n}_{1}\in\boldsymbol{R}^{\frac{T}{r} \times T}$ and $\boldsymbol{W}^{n}_{2} \in \boldsymbol{R}^{T \times {\frac{T}{r}}}$ are trainable parameter matrices, and optional parameter $r$ is used to control the model complexity, $f(\cdot)$ is a Heaviside step function that is same as in Eq.\ref{eq:SNN layer}, and $d_{t h}$ is the attention threshold. We use the score vector to train a complete network at the training stage. As an optional operation, at the inference stage, we discard the irrelevant frames which are lower than $d_{t h}$, and set the attention score of the other frames to 1.

Finally, we use $\boldsymbol{d}^{n-1}$ as the input score vector, and the final input at $t$th timestep is
\begin{equation}
\widetilde{\boldsymbol{X}}^{t,n-1}=\boldsymbol{d}^{n-1}_{t}\boldsymbol{X}^{t,n-1}
\label{Eq:attention_eq3}
\end{equation}
where $\widetilde{\boldsymbol{X}}^{t, n-1}\in\boldsymbol{R}^{L\times B\times C}$ is $\boldsymbol{X}^{t,n-1}$ with attention score at $t$th timestep in Eq.\ref{eq:SNN layer}. Then, the membrane potential behaviors of a TA-LIF and TA-LIAF layer follow
\begin{equation}
    \boldsymbol{U}^{t, n}=\boldsymbol{H}^{t-1, n}+g\left(\boldsymbol{W}^{n}, \widetilde{\boldsymbol{X}}^{t, n-1}\right). \\
\end{equation}

The excitation step maps the statistical vector $\boldsymbol{z}$ to a set of temporal-wise input scores. In this regard, the TA module can be deemed as a self-attention function. The main difference is that statistical vectors in the frame-based representation directly correlate with the number of events.

\section{Experiments}\label{Experiment}

\subsection{Experimental Setup}
\textbf{Datasets}. We perform experiments on three kinds of classification datasets, which are all event datasets but are obtained in different ways. The first is DVS128 Gesture\cite{Gesture_Recognition_System}, which is a gesture recognition dataset capture by DVS cameras. The second is CIFAR10-DVS\cite{Cifar10-DVS}, which is an event-based image classification dataset convert from the static dataset by scanning each sample in front of DVS cameras. The last one is the Spoken Heidelberg Digits (SHD)\cite{SHD_dataset}, which is an audio classification dataset convert from audio by software simulation.

\textbf{Learning}. Table \ref{Table:Unification for comparison} lists details for experiments like learning algorithm, loss function, etc. We use the Adam optimizer \cite{DBLP:journals/corr/KingmaB14} for accelerating the training process and employ some standard training techniques of deep learning, such as batch normalization, dropout, etc, and the corresponding hyper-parameters and SNN hyper-parameters are shown in Table \ref{Table:hyper-parameters_setting}. The network structures of the three tasks are shown in Table \ref{Table:Network_Arch}, and we adopt the same network structure for LIF-SNN and LIAF-SNN in each dataset. 

\textbf{RCS Method}. Leave out the time consumption in hardware, event-based system latency $t_{lat}$ only hinges on $dt$ and $T$, i.e., $dt \times T$. Inspired by the random temporal cropping during video recognition method\cite{LTC}, we apply similar data augmentation at the training stage, which is termed as RCS, i.e., select a random $t_{0}$ (see Fig.\ref{fig:Event_data_fig}) as the starting point and aggregate consecutive frames. At the test time, we adopt a voting mechanism by following \cite{LTC}, that is, for the given $dt$ and $T$, an event stream is divided into consecutive 10-crops and the length of each one is $t_{lat}$, and the final test result is obtained by accumulating the results of all the individual crops. If the number of frames is less than $10 \times t_{lat}$, we adopt overlap methods in \cite{Gesture_Loihi}, e.g., using 2-crops of 30ms and $t_{lat} =20ms$, the crops will cover partially overlapped ranges as [$0ms$; $20ms$] and [$10ms$; $30ms$].

\begin{table}[t]
    \centering
    \caption{Unification for comparison. Our network implements on the Pytorch\cite{pytorch} framework.}
    \setlength{\tabcolsep}{0.8mm}
    {
        \begin{tabular}{|c|c|}
        \hline
        \multirow{2}{*}{Dataset} & CIFAR10-DVS \& DVS128 Gesture \\ & \& SHD Dataset\\
        \hline
        Representation  & Tunable Frames    \\
        Output Latency &  $dt \times T$          \\
        Learning Algorithm  & STBP\cite{STBP_Alogorithm} \& BPTT \\
        Loss Function  & Rate Coding\cite{Compare_SNN_and_RNN}     \\
        Network Structure  & CNN-based SNN\cite{Compare_SNN_and_RNN}                \\
        \hline
    \end{tabular}
    }
    \label{Table:Unification for comparison}
\end{table}

\begin{table}[t]
    \centering
    \caption{Hyper-parameter setting.}
    \setlength{\tabcolsep}{0.6mm}
    {
    \begin{tabular}{|c|c|c|c|}
        \hline
        Hyper parameter & DVS128 Gesture & CIFAR10-DVS & SHD \\
        \hline
        Max Epoch & 100 & 150 & 100\\
        Batch Size & 36 & 64& 256\\
        Learning Rate  &$1{e}^{-4}$ & $1{e}^{-3}$& $1{e}^{-3}$\\
        $u_{t h}$ & 0.3 & 0.3& 0.3\\
        $e^{-\frac{d t}{\tau}}$ & 0.3 & 0.3 & 0.3\\
        $r$ & 16 & 5 & 5\\
        \hline
    \end{tabular}
    }
    \label{Table:hyper-parameters_setting}
\end{table}{}

\subsection{Gesture Recognition}\label{Gesture_exper}

\begin{table}[t]
    \centering
    \caption{Network structure. MP4-max pooling is $4\times4$, $n$C3-Conv is $3\times3$ and has $n$ output feature maps, AP2-average pooling is $2\times2$, $n$FC-Linear layer has $n$ output feature maps.}
    \setlength{\tabcolsep}{4.5mm}{

    \begin{tabular}{|c|c|}
        \hline
        Dataset & Network Structure\\
        \hline
        \multirow{2}{*}{DVS128 Gesture} & Input-MP4-64C3-128C3-\\
                                        & AP2-128C3-AP2-256FC-11 \\\hline
        \multirow{3}{*}{CIFAR10-DVS}   & Input-32C3-AP2-64C3-\\
         & AP2-128C3-AP2-256C3-\\
         & AP2-512C3-AP4-256FC-10\\\hline
        
        SHD Dataset & Input-128FC-128FC-20\\
        \hline
    \end{tabular}

       }
    \label{Table:Network_Arch}
\end{table}{}

\begin{figure}[t]
\centering
\includegraphics[scale=0.52]{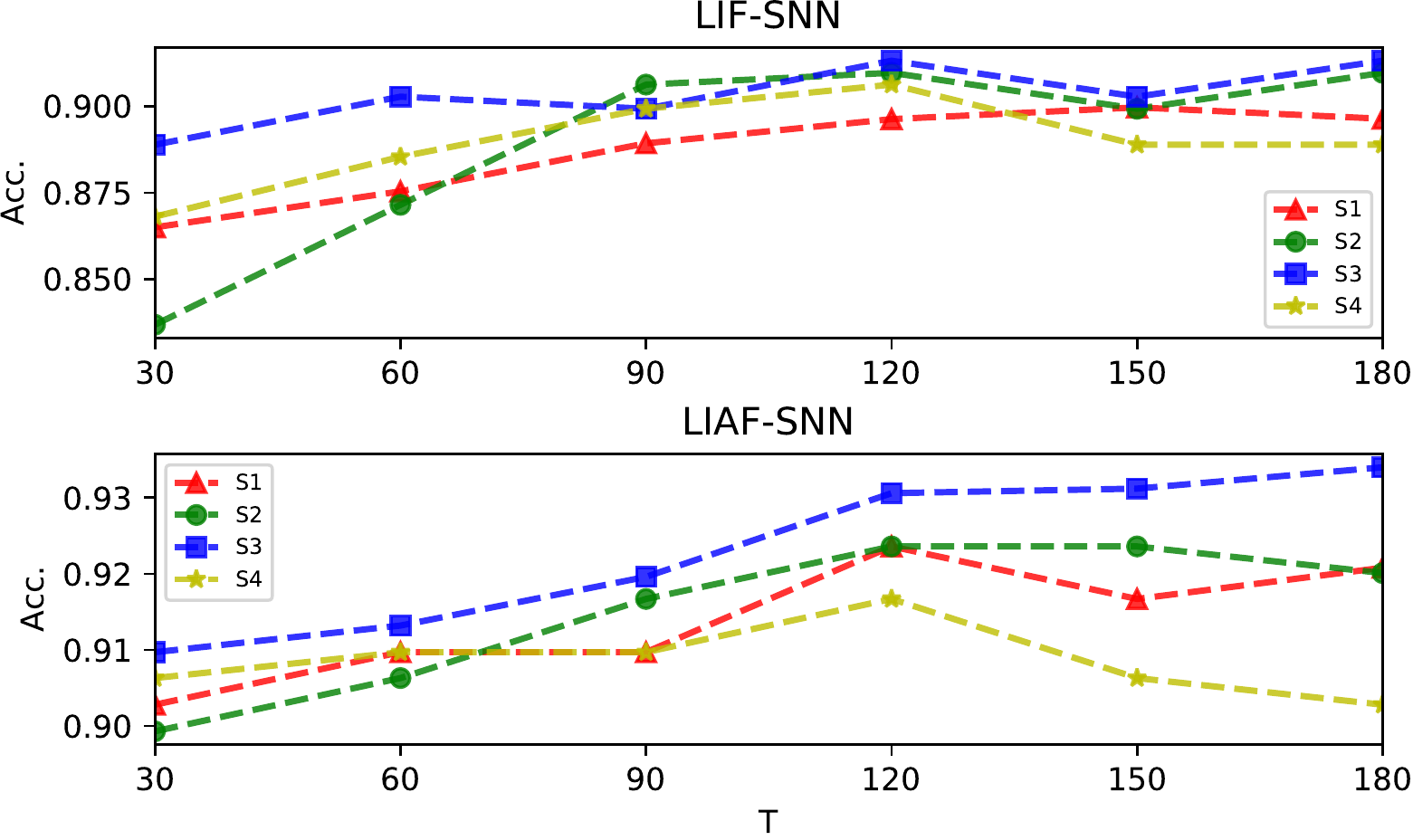}
\caption{Ablation study of different TA positions based on DVS128 Gesture in (a) LIF-SNN, and (b) LIAF-SNN. S1, pure SNNs without TA; S2, insert TA only at the input layer; S3, insert TA only at the depth layers; S4, insert TA to the whole network.}
\label{fig:Figure3}
\end{figure}

The IBM DVS128 Gesture\cite{Gesture_Recognition_System} is an event-based gesture recognition dataset, which has the TR in \textmu s level and $128 \times 128$ spatial resolution. It records 1342 samples of 11 gestures, and each gesture has an average duration of 6 seconds. Note that DVS128 Gesture has two kinds of categories which are 10 and 11 classes, and we select the latter setting that is harder.

\begin{figure*}[t]
\centering
\includegraphics[width=16cm,height=5cm]{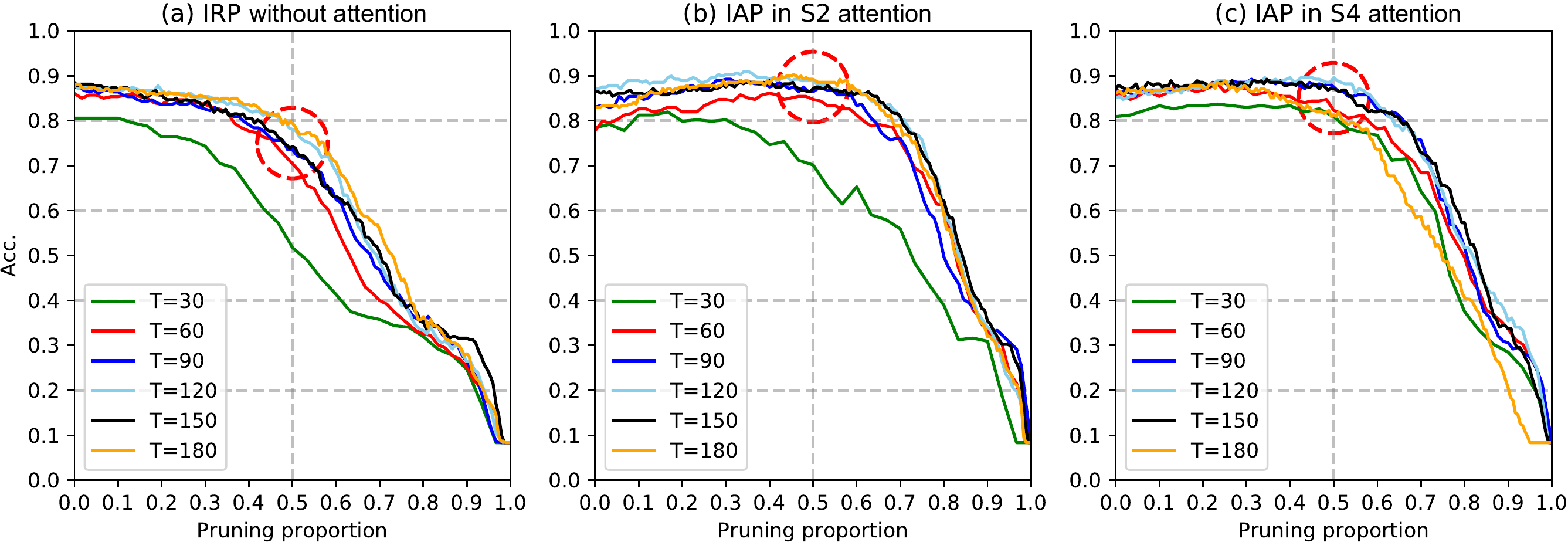}
\caption{Experiments of IAP.  We fixed $dt=1ms$, varying simulation timestep $T\in\left\{30,60,90,120,150,180\right\}$ and pruning proportion on DVS128 Gesture. From left to right (a) IRP without attention. (b) TA module only used in the input layer with IAP. (c) TA module used in the whole network with IAP. As shown by the dotted circle, when the pruning proportion is 0.5, most of the IAP results still maintain high accuracy around 89\%, but the IRP accuracy results decrease to around 78\%.}
\label{fig:figure4}
\end{figure*}

\textbf{Ablation Study of Different TA Positions}. The position to insert TA is important, and to evaluate its influence, we design an ablation study of different TA positions, which consist of four: \textbf{S1, pure SNNs without TA; S2, insert TA only at the input layer; S3, insert TA only at the depth layers (whole network except the input layer); S4, insert TA to the whole network}. Perceptually, smooth interactions in real gesture recognition tasks require systems to respond within 100-200 ms\cite{Gesture_Recognition_System}. Based on this requirement, we set $T\in\left\{30,60,90,120,150,180\right\}$ with $dt=1ms$. The impact of the TA positions and simulation timestep $T$ on gesture recognition are shown in Fig.\ref{fig:Figure3}. For TA positions, we observe that using the S2 (green) or S3 (blue) independently can improve performance in most cases, and S3 achieves the best accuracy. But combining S2 and S3 (i.e., S4, yellow) leads to unstable results, and the accuracy keeps going down when $T$ grows bigger. Besides, when T falls in the first half range, improving T can improve the accuracy slightly, but further enlarging T ($T>120$) will be helpless.

\textbf{Experiments of IAP}. Inspired by the characteristic of event-triggered computation in LIF-SNNs, we discard the frames with lower attention scores at the inference stage (see Eq.\ref{Eq:traning_and_inference}) for power-saving and term this method as IAP. It is worth noting that the attention mechanism brings the possibility for the discard operation. To evaluate the effects of input pruning, we set IAP on S2 and S4 since each frame has an attention score in these two cases. To make an intuitive comparison, without attention, we choose a simple \emph{input random pruning (IRP)} to achieve the same level of power consumption as the baseline. The accuracies of IRP in Fig.\ref{fig:figure4} (a) appear approximative monotone decreasing with the increase of the pruning proportion. However, for Fig.\ref{fig:figure4} (b) and (c), the accuracies do not decrease as the pruning proportion increases at the first half of the pruning proportion. Detailedly, as shown by the dotted circle in Fig.\ref{fig:figure4}, when the pruning proportion is 0.5, most of the IAP still maintain high accuracies around 89\%, but the IRP accuracies decrease to around 78\%. Moreover, as shown in Table \ref{Table:compare_flot}, the best pruning proportion of the IAP relates to the simulation timestep $T$, and we can get similar or even better performance with almost only half the power and a low-latency (30ms to 180ms) with the TA module compared with using full input. 

\begin{table}[t]
    \centering
    \caption{Influence of IAP in variant T on parameters, accuracy, and FLOPs. ``Param.'' means the ratio of increased parameters of TA, ``Best Pro.'' and ``Acc.'' reflect the best pruning proportion to keep the accuracy. ``FLOPs'' means the floating point operations.} 
    \setlength{\tabcolsep}{1.5mm}
    {
    \begin{tabular}{|c|c|c|c|c|c|}
        \hline
         \multirow{2}{*}{IAP} & \multirow{2}{*}{T}& Param.& \multirow{2}{*}{Best Pro.} & \multirow{2}{*}{Acc.(\%)} & FLOPs  \\
         & & (\%$\uparrow$) &  &  & (\%$\downarrow$)  \\ \hline
         \multirow{6}{*}{S2} & 30 & +0.004 & 0.17 & 81.95(-1.73) & -16.67   \\ 
          &60 & +0.018 & 0.40 & 86.11(-1.04) & -40.00   \\ 
          &90 & +0.043 & 0.30 & 89.24(-1.39) & -30.00   \\ 
          &120 & +0.078 & 0.35 & 90.99(+0.02) & -35.00   \\ 
          &150 & +0.123 & 0.40 & 89.89(-0.04) & -40.00   \\ 
          &180 & +0.179 & 0.40 & 91.28(+0.31) & -40.00   \\ 
          \hline
          \multirow{6}{*}{S4}&  30 & +0.020 & 0.34 & 83.33(-3.48) & -33.33 \\ 
          &  60 & +0.091 & 0.24 & 88.20(-0.34) & -23.33 \\
          &  90 & +0.214 & 0.32 & 89.24(-0.69) & -31.11 \\
          & 120 & +0.389 & 0.50 & \textbf{90.58(-0.05)} & \textbf{-50.00} \\
          & 150 & +0.615 & 0.35 & 89.24(+0.35) & -34.67 \\
          & 180 & +0.893 & 0.35 & 88.89(+0.00) & -35.00 \\
          \hline
    \end{tabular}
    }
    \label{Table:compare_flot}
\end{table}

\begin{table}[t]
    \centering
    \caption{Ablation study of RCS and TA-SNNs. We use the S3 strategy to test multiple-scale TRs with $T=60$.} 
    \setlength{\tabcolsep}{0.5mm}
    {
    \begin{tabular}{|c|c|c|c|c|c|}
        \hline
        \multirow{2}{*}{Model} & \multirow{2}{*}{$dt$} & SNN(\%) & TA-SNN & SNN & TA-SNN\\
        & & \cite{Compare_SNN_and_RNN} & (\%) & (RCS)(\%) & (RCS)(\%)\\
        \hline
        \multirow{6}{*}{LIF} & 1ms & 71.53 & 73.25 & 87.15 & \textbf{90.28(+18.75)} \\
         & 5ms & 87.15 & 89.24 & 90.63 & 93.40 \\
         & 10ms & 91.67 & 93.40 & 93.40 & 94.79 \\
         & 15ms & 93.05 & 95.49 & 92.36 & 95.49 \\
         & 20ms & 92.71 & 94.44 & 91.32 & 94.79 \\
         & 25ms & 93.40 & 95.14 & 91.67 & 95.48 \\
        \hline
         \multirow{6}{*}{LIAF}& 1ms & 72.59 & 74.31 & 90.97 & \textbf{91.32(+18.73)} \\
         & 5ms & 88.20 & 89.93 & 93.06 & 94.10 \\
         & 10ms & 93.75 & 95.14 & 93.75 & 94.79 \\
         & 15ms & 95.14 & 96.88 & 94.10 & 94.79 \\
         & 20ms & 95.84 & 97.57 & 94.10 & 95.14 \\
         & 25ms & 96.18 & \textbf{98.61} & 94.44 & 94.79 \\
        \hline
    \end{tabular}
    }
    \label{Table:TA_SNN_exper}
\end{table}

\textbf{Ablation Study of RCS and TA}. To investigate the influence of RCS and TA, we conduct several ablation studies in Table \ref{Table:TA_SNN_exper}. Because of the stability of the S3 strategy, it will be used for all the rest of experiments in this paper. For the comprehensiveness of the studies, we set multiple-scale $dt\in\left\{1,5,10,15,20,25\right\}$ with fixed $T=60$. First, we show the effect of the TA and RCS method individually based on benchmark SNN results \cite{Compare_SNN_and_RNN}. We observe that TA works in all conditions, and RCS makes a great improvement of accuracy when $dt$ is small, but the effect is weakened when $dt$ is bigger. Next, we apply those methods on LIF and LIAF, and get variant results. For LIF, RCS and TA can work together very well with an accuracy of 95.49\%. For LIAF, RCS has a negative influence when $dt$ is bigger ($dt\in\left\{15,20,25\right\}$), and we reports the best accuracy of 98.61\% without RCS.

\subsection{Event-based Image Classification}
\begin{figure}[t]
\centering
\includegraphics[width=8cm,height=3.5cm]{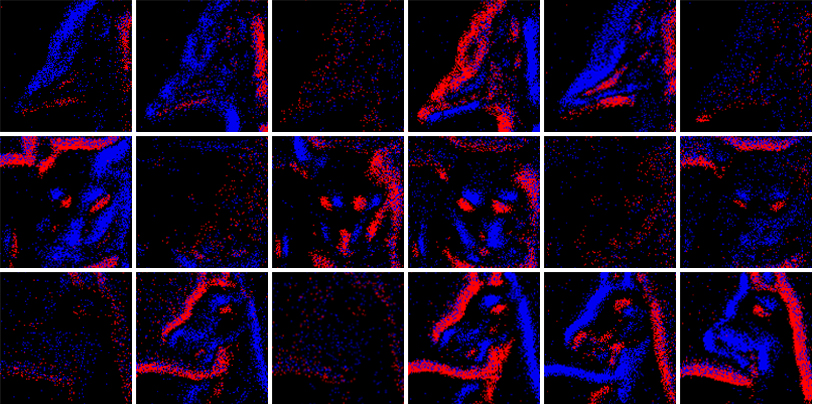}
\caption{Examples of consecutive frames in CIFAR10-DVS with $dt=10ms$. The movement of images in CIFAR10 is designed by fixed trajectory, and the distance of spatial movement is restricted. Thus, frames at different timestep are similar, and the temporal feature is not the dominant information\cite{rethink_ann_snn}.}
\label{fig:cat_fig}
\end{figure}

CIFAR10-DVS\cite{Cifar10-DVS} is an event-based dataset converted from CIFAR10 by scanning each image with repeated closed-loop movement in front of a DVS. CIFAR10-DVS includes 1000 samples for each category in CIFAR10, and there are in total 10,000 samples, with each one having a duration of 300ms. The temporal and spatial resolutions are \textmu s and $128 \times 128$, respectively. Unlike gesture recognition, the temporal feature in CIFAR10-DVS may not be dominant\cite{rethink_ann_snn}. Fig.\ref{fig:cat_fig} gives examples in CIFAR10-DVS, which can be observed that the temporal correlation between different frames is not obvious. Based on the above analysis, we select moderate parameters that are $T=10$ and $dt = 10ms$. As shown in Table \ref{Table:Accuracy of CIFAR10-DVS Dataset}, in these experiments, both RCS and TA-SNNs can improve accuracy. 


\begin{table}[t]
    \centering
    \caption{Ablation experiments on $dt=10ms$ and $T=10$ in the CIFAR10-DVS dataset with the S3 strategy.}
    \setlength{\tabcolsep}{1.0mm}
    {
    \begin{tabular}{|c|c|c|c|c|c|}
        \hline
        \multicolumn{3}{|c|}{LIF}  & \multicolumn{3}{c|}{LIAF} \\ \cline{1-6}
          \multirow{2}{*}{SNN}  &SNN  & TA-SNN   &\multirow{2}{*}{SNN}    &SNN & TA-SNN\\  
            &(RCS)  & (RCS)   &   &(RCS)  & (RCS)\\  \hline
          54.70\% & 66.60\% & \textbf{71.10}\%  & 69.40\% & 70.97\% & \textbf{72.00}\%    \\ \hline
    \end{tabular}
    }
    \label{Table:Accuracy of CIFAR10-DVS Dataset}
\end{table}

\begin{table}[t]
    \centering
    \caption{Experiments on TA-SNNs in SHD dataset with the S3 strategy. For comparison, we fixed $t_{lat}$ here and adopt the same network structure with \cite{SHD_2}. Since shorter samples will pad with zeros, the RCS method cannot be used here.}
    \setlength{\tabcolsep}{1.6mm}
    {
    \begin{tabular}{|c|c|c|c|c|c|}
        \hline
        \multirow{3}{*}{$dt$ }  & \multirow{3}{*}{$T$} & \multicolumn{2}{c|}{LIF}  &\multicolumn{2}{c|}{LIAF} \\ \cline{3-6}
                         &  &\multirow{2}{*}{SNN(\%)} & TA-SNN  &\multirow{2}{*}{SNN(\%)} & TA-SNN\\  
                          &  & & (\%)  & & (\%)\\  \hline
        \multirow{3}{*}{4ms}  & 50  & 54.33 & 57.77  & 58.75 &   61.23    \\ 
                              & 150 & 74.16 & 85.91  & 75.04 &   82.24    \\ 
                              & 250 & 75.88 & 84.50  & 75.49 &   81.45  \\  \hline
        \multirow{3}{*}{10ms} & 20  & 79.42 & 84.76  & 78.40 &   79.24   \\ 
                              & 60  & 77.52 & 86.71  & 87.68 &   86.32   \\ 
                              & 100 & 81.45 & 86.66  & 84.54 &   88.21     \\  \hline
        \multirow{2}{*}{60ms} & 10  & 86.79 & 87.59  & 89.05 &  \textbf{91.08}   \\ 
                              & 15  & 85.87 & 86.88  & 86.35 &   89.80    \\ \hline
    \end{tabular}
    }
    \label{Table:Accuracy of SHD Dataset}
\end{table}

\subsection{Audio Classification}
\begin{figure}[t]
\centering
\includegraphics[width=8cm,height=3.95cm]{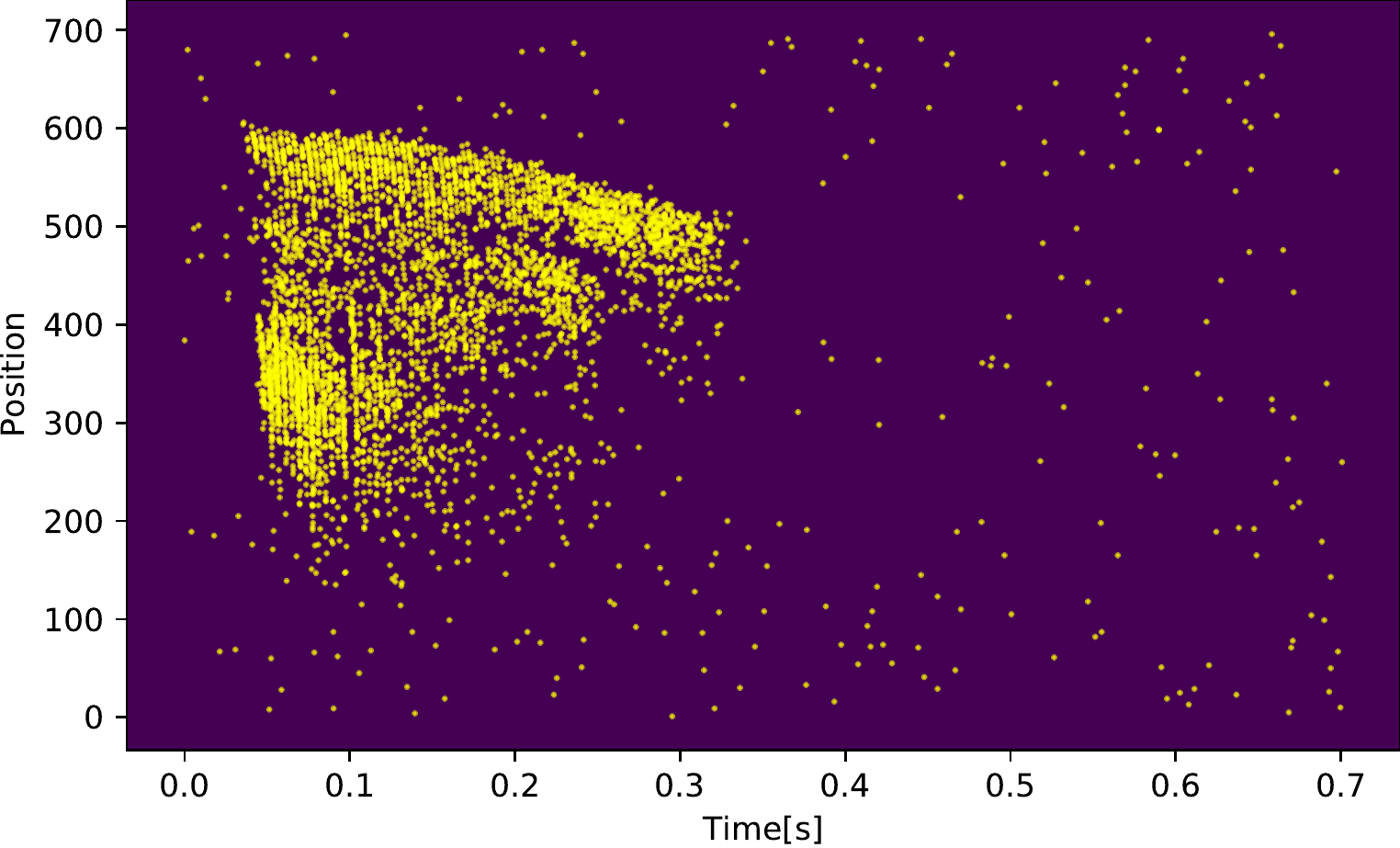}
\caption{An audio example from the SHD dataset. The data contents of audio have no periodicity, and this is essentially different with natural gestures and the periodic movement of the image.}
\label{fig:SHD}
\end{figure}

The SHD dataset\cite{SHD_dataset} is a large spike-based audio classification task that contains 10420 audio samples of spoken digits ranging from zero to nine in English and German languages. A biologically inspired model\cite{SHD_dataset} is used to convert the audio signal into a spike stream, and the data duration ranges from 0.24s to 1.17s. Unlike the four-dimensional event stream generated by the DVS camera, the audio spike stream has only two dimensions, i.e., time and position. As shown in Fig.\ref{fig:SHD}, the resolution of time dimension is \textmu s level, and the position ranges from 0 to 699. We adopt the same data preprocess method in \cite{SHD_2}, i.e., all samples are fit within a 1s window, where shorter samples are padded with zeros, and longer samples are cut. We use $t_{lat} \in\left\{200,600,900,1000\right\}$ with different $dt$ based on S3 strategy. Results in Table \ref{Table:Accuracy of SHD Dataset} demonstrate that TA-SNNs always work in all kinds of parameter combinations, and the bigger $t_{lat}$ is, the higher accuracy we could have. To verify the effectiveness of the TA module, we also insert it into an extra analog-based SNN, i.e., ReLU SRNN\cite{SHD_2}, and report 90.02\% accuracy, which is higher than 88.93\% in \cite{SHD_2}.

\begin{table}[t]
\centering
\caption{Accuracy of models for the DVS128 Gesture, CIFAR10-DVS and SHD Dataset.}
\setlength{\tabcolsep}{0.1mm}
{
\begin{tabular}{|c|c|c|c|}
\hline
Task & Proposals & Methods   & Acc.(\%)\\    \hline
        &Amir \etal~2017\cite{Gesture_Recognition_System}  & 12 layers CNN & 94.59        \\ 
        &Shrestha \etal~2018\cite{Slayer}  & Slayer & 93.64       \\ 
        &Wu \etal~2020\cite{LIAF}    & LIAF-Net    &       97.56      \\   
DVS 128 &Kugele \etal~2020\cite{efficient_snn}   & DenseNet SNN &        95.56               \\ 
Gesture &Massa \etal~2020\cite{Gesture_Loihi}  &  SNN on Loihi &   89.64         \\  
&Zheng \etal~2020\cite{Going_Deeper_SNN}    & ResNet17 SNN        & 96.87          \\
&Khoei \etal~2020\cite{SpArNet}    & SpArNet       & 95.10          \\
 &He \etal~2020\cite{Compare_SNN_and_RNN}    & LIF-Net  &       93.40 \\ 
    & \textbf{This work (SOTA)} &  TA-SNN  &     \textbf{98.61}  \\\hline
    
&Wu et.al. 2018\cite{Direct_training}    & NeuNorm SNN  &  60.50\\ 
&Ramesh \etal~2019\cite{DART} &  DART   &   65.78  \\ 
CIFAR10&Wu \etal~2020\cite{LIAF}    & LIAF-Net   & 70.40     \\ 
-DVS&Kugele \etal~2020\cite{efficient_snn}  &  SR-ANN  & 66.75   \\ 
&Zheng \etal~2020\cite{Going_Deeper_SNN}  & ResNet19 SNN &   67.80 \\ 
&\textbf{This work (SOTA)} & TA-SNN & \textbf{72.00} \\\hline    

&Cramer \etal~2020\cite{SHD_dataset} & LIF RSNN &        71.40 \\ 
SHD&Yin \etal~2020\cite{SHD_2} &  RELU SRNN   &         88.93       \\
Dataset&Zenke \etal~2021\cite{2021The} &  SG-based SNN   &  84.00        \\
&\textbf{This work (SOTA)}    & TA-SNN  & \textbf{91.08}   \\\hline 
\end{tabular}
}
\label{Table:Compare with other SNN model in three kinds of tasks.}
\end{table}

\subsection{Comparison with Prior Works}

We compare our best results of the proposed TA-SNN against various of prior works for event-based data, such as CNN method\cite{Gesture_Recognition_System,Slayer}, spike-based SNNs\cite{efficient_snn,Gesture_Loihi,Going_Deeper_SNN,Direct_training,SHD_dataset,Compare_SNN_and_RNN,2021The}, and analog-based SNNs\cite{LIAF,SpArNet,SHD_2}, etc. As shown in Table \ref{Table:Compare with other SNN model in three kinds of tasks.}, our TA-SNN models achieve the SOTA in various datasets, and the performance of spike-based SNNs and analog-based SNNs have been improved by inserting the TA module. Moreover, comparing with the original SNNs, the number of parameters in TA-SNNs almost has no increase. \textbf{From the above comparisons, it can be seen that the TA module can help SNN to achieve higher performance with less cost in various tasks, thereby, the TA module will contribute a lot to promote SNNs to practical applications.}

\section{Discussion}

\textbf{TA position}. Different TA positions in SNNs have different effects on performance, and the TA module inserted on depth layers (i.e., S3) works better (see Fig.\ref{fig:Figure3}). This phenomenon is similar to the SE block used in the channel domain, where the SE in deeper layers is slightly better than that in lower layers\cite{SE_PAMI}. Compared with pure SNNs, TA-SNNs based on the S3 can always enhance the network's ability to extract spatio-temporal features. 

\textbf{Adaptability of TA module}. One of the most valuable points in the spike-based SNNs is the event-triggered computation feature\cite{DVS_Survey}. However, keeping the event-triggered characteristic also brings difficulty in training since the spike activity is hard to differentiate. Although the STBP algorithm, which can solve the differentiability issue to some extend, appears to be barely satisfactory in deep SNNs, comfortingly, our TA module can improve the accuracy of spike-based SNNs. Moreover, instead of utilizing full input frames, our TA module also brings interesting and important IAP that can get similar or even better performance with only half of the input frames and low latency (30ms to 180ms). This achievement may magnify the advantage of power-saving and exhibit the potential of network performance improvement in deep spike-based SNNs. On the other hand, the TA module also works in analog-based SNNs, which give up the event-triggered characteristic but keep the dynamic characteristics of a biological neuron, and all SOTA results are obtained in this way, however, more power might be needed in return.

\textbf{Influence of RCS method}. Prior works mostly used $t_{0}=0$ as the starting point in training, while our RCS method selects a random $t_{0}$ (see Fig.\ref{fig:Event_data_fig}). For the RCS, there is a basic precondition that the content of event streams should have inherent cycles of repetition. Both gesture action in DVS128 Gesture and repeated movement of the image in CIFAR10-DVS satisfy this precondition. As shown in Table \ref{Table:TA_SNN_exper}, RCS works better under short-term $dt$. However, using RCS with long-term $dt$ will reduce accuracy, e.g., the accuracy of analog-based SNNs reduces from 98.61\% to 94.79\% with $dt=25ms$. The possible reason is that choosing a long-term aggregation window will destroy the inherent periodicity. In SHD, selecting a random $t_{0}$ may likely cause all the input data to be 0 for a shorter sample, thus RCS does not work here either. 

\textbf{TR Analysis}. TR is a crucial hyper-parameter for frame-based representation. Current methods usually adopt a long-term TR to make sure SNR is sufficient in each frame. It is indeed useful since all SOTA results in this work are obtained in the long-term TR. However, long-term TR will dilute the advantages of the asynchronicity and sparsity of event streams and increase the output latency. Short-term TR brings high-rate frames that are friendly to high-speed object tracking and low-latency interaction, but the produced low SNR is an intractable issue for getting satisfactory performance. By using RCS and TA, this issue is greatly relaxed in our work. Firstly, the RCS method significantly strengthens short-term TR's advantages, i.e., the smaller the TR is, the more optional training data we can organize. Secondly, the TA enhances the ability of SNNs to effectively extract spatio-temporal features (see Table \ref{Table:TA_SNN_exper}). Moreover, IAP with short-term TR in spike-based TA-SNN can keep or improve task accuracy. These experiments imply that event streams with short-term TR have a great potential to solve the real-time scenarios with considerable accuracy. Last but not least, the selection of TR also depends on the inherent trait of different tasks or datasets. In our experiments, DVS128 Gesture naturally has affluent repeatability, so short-term TR can obtain an acceptable result. Meanwhile, CIFAR10-DVS has poor temporal features and SHD sample almost has no repeatability, so short-term TR is meaningless for them.

\section{Conclusions}

In this paper, we innovatively integrate the temporal attention mechanism into SNNs and propose the TA-SNNs that can deal with the event streams more effectively and efficiently than the pure LIF-SNNs while preserving SNNs' event-triggered feature. Additionally, attention-score-based input pruning technology is used in the inference process, which surprisingly doesn't cause a significant accuracy loss but saves a large amount of computation. We also propose the RCS method and investigate the performance of TA-SNNs on various datasets in different TRs. The experiment results are provided using TA-SNNs and RCS, and achieve SOTA results in DVS128 Gesture (98.61\%), CIFAR10-DVS (72.00\%), and SHD (91.08\%), verifying the effectiveness of these methods.

We believe that this method will greatly expand people's imagination of SNNs, guide more advanced deep learning technology into SNNs research, and open up the way for the applications of SNNs. In addition, in future work, this method will also help SNNs to get better performance on hardware. The sparse event-triggered characteristics of SNNs are kept by TA-SNNs, which will be of great significance to improve the performance on the SNNs accelerators.

\section*{Acknowledgment}
This work was partially supported partially by National Key R\&D Program of China (2018AAA0102600, 2018YEF0200200), National Natural Science Foundation of China (61876215, 12002254), and Beijing Academy of Artificial Intelligence (BAAI), and the open project of Zhejiang Laboratory, and a grant from the Institute for Guo Qiang of Tsinghua University, and Peng Cheng Laboratory.

{\small
\bibliographystyle{ieee_fullname}
\bibliography{egbib}
}

\end{document}